\documentclass{article}

\usepackage{ijcai19}

\usepackage{times}
\usepackage{tabularx}
\usepackage{soul}
\usepackage{url}
\usepackage[font=small,skip=0pt]{caption}
\usepackage[colorlinks=false]{hyperref}
\usepackage[utf8]{inputenc}
\usepackage{graphicx}
\usepackage{amsmath}
\usepackage{booktabs}
\usepackage{subcaption}
\usepackage{multirow}
\usepackage[algo2e]{algorithm2e}
\usepackage{algorithm}
\usepackage[noend]{algpseudocode}
\hypersetup{%
    ,citecolor=black
    }

\title{Harnessing the Vulnerability of Latent Layers in Adversarially Trained Models}
\author{
Nupur Kumari$^1$\thanks{ Authors contributed equally}
\and
Mayank Singh$^1$$^{*}$\and
Abhishek Sinha$^1$$^{*}$\and
Harshitha Machiraju$^2$\and
Balaji Krishnamurthy$^1$\And
Vineeth N Balasubramanian$^2$
\affiliations
$^1$Adobe Inc,Noida\\
$^2$IIT Hyderabad
\emails
\{ nupkumar,msingh,abhsinha,kbalaji \} @adobe.com\\ \{ ee14btech11011,vineethnb \} @iith.ac.in 
}

\begin{document}

\maketitle

\begin{abstract}
Neural networks are vulnerable to adversarial attacks - small visually imperceptible crafted noise which when added to the input drastically changes the output. The most effective method of defending against adversarial attacks is to use the methodology of adversarial training. We analyze the adversarially trained robust models to study their vulnerability against adversarial attacks at the level of the latent layers. Our analysis reveals that contrary to the input layer which is robust to adversarial attack, the latent layer of these robust models are highly susceptible to adversarial perturbations of small magnitude. Leveraging this information, we introduce a new technique Latent Adversarial Training (LAT) which comprises of fine-tuning the adversarially trained models to ensure the robustness at the feature layers. We also propose Latent Attack (LA), a novel algorithm for constructing adversarial examples. LAT results in a minor improvement in test accuracy and leads to a state-of-the-art adversarial accuracy against the universal first-order adversarial PGD attack which is shown for the MNIST, CIFAR-10, CIFAR-100, SVHN and Restricted ImageNet datasets.
\end{abstract}

\section{Introduction}
Deep Neural Networks have achieved state of the art performance in  several Computer Vision tasks \cite{he2016deep,krizhevsky2012imagenet}. However, recently it has been shown to be extremely vulnerable to adversarial perturbations.  These small, carefully calibrated perturbations when added to the input lead to a significant change in the network's prediction \cite{szegedy2013intriguing}. The existence of adversarial examples pose a severe security threat to the practical deployment of deep learning models, particularly, in safety-critical systems \cite{reviewpaper2}.\\
\begin{figure}[t]
\setlength{\belowcaptionskip}{-10pt}
\begin{center}
    \includegraphics[width=3.4in,height=1.8in]{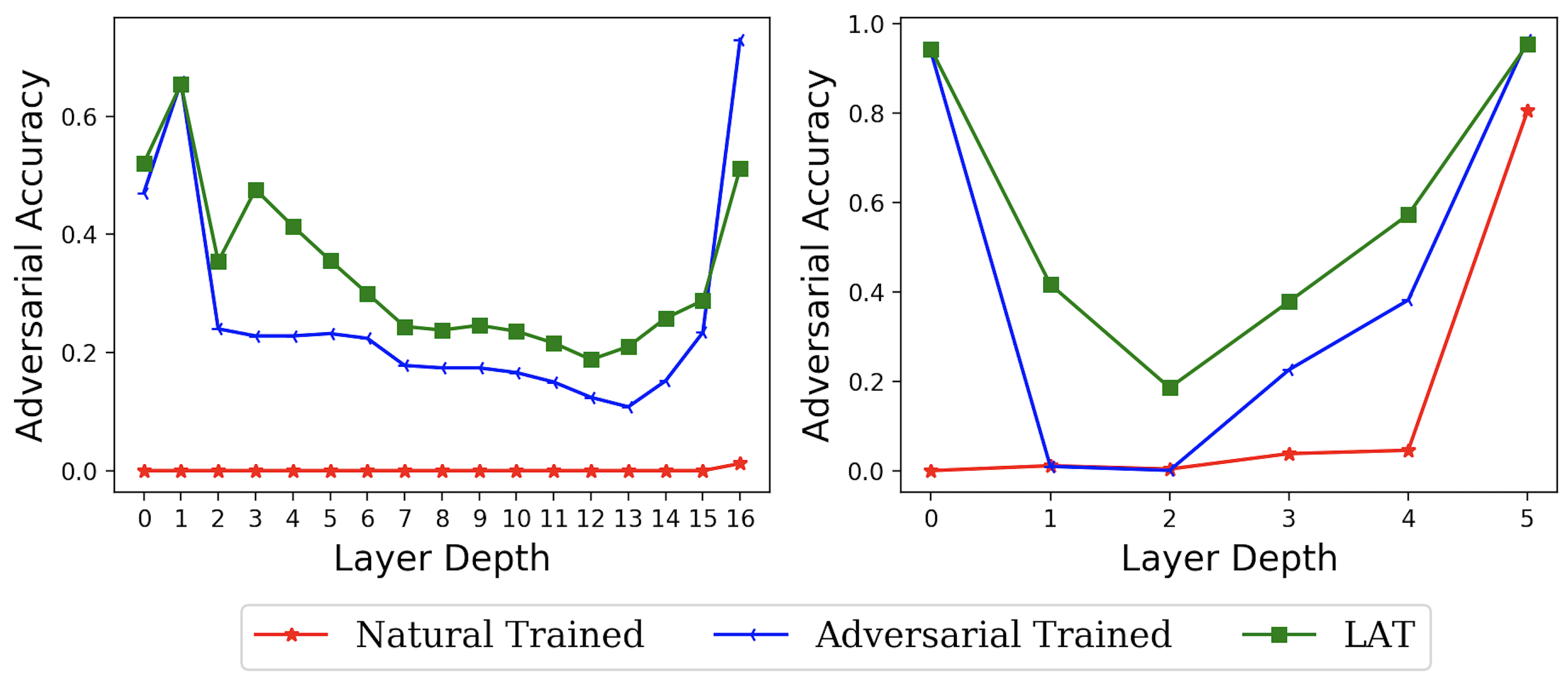} 
     \label{layer_mnist_cifar}
    \caption{Adversarial accuracy of latent layers for different models on CIFAR-10 and MNIST}
    \label{latent_acc}
\end{center}
\end{figure}
Since the advent of adversarial perturbations, there has been extensive work in the area of crafting new adversarial attacks \cite{attack2017pgd,moosavi2017universal,carlini2017towards}. At the same time, several methods have been proposed to protect models from these attacks (adversarial defense)\cite{goodfellow2014explaining,attack2017pgd,tramer2017ensemble}. Nonetheless, many of these defense strategies are continually defeated by new attacks. \cite{obfuscate,carlini2017adversarial,attack2017pgd}. In order to better compare the defense strategies, recent methods try to provide robustness guarantees by formally proving that no perturbation smaller than a given $l_p (\text{where } p \in [0,\infty])$bound can fool their network \cite{raghunathan2018certified,tsuzuku2018lipschitz,weng2018towards,provable_certificate4,wong2017provable}. Also some work has been done by using the Lipschitz constant as a measure of robustness and improving upon it \cite{szegedy2013intriguing,cisse2017parseval,tsuzuku2018lipschitz}.

Despite the efforts, the adversarial defense methods still fail to provide a significant robustness guarantee for appropriate $l_p$ bounds (in terms of accuracy over adversarial examples) for large datasets like CIFAR-10, CIFAR-100, ImageNet\cite{russakovsky2015imagenet}. Enhancing the robustness of models for these datasets is still an open challenge.

In this paper, we analyze the models trained using an adversarial defense methodology \cite{attack2017pgd} and find that while these models show robustness at the input layer, the latent layers are still highly vulnerable to adversarial attacks as shown in Fig \ref{latent_acc}. We utilize this property to introduce a new technique (LAT) of further fine-tuning the adversarially trained model. We find that improving the robustness of the models at the latent layer boosts the adversarial accuracy of the entire model. We observe that LAT improves the adversarial robustness by ($\sim 4-6\%$) and test accuracy by ($\sim1\%$) for CIFAR-10 and CIFAR-100 datasets. 

Our main contributions in this paper are the following: 
\begin{itemize}
    \item We study the robustness of latent layers of networks in terms of adversarial accuracy and Lipschitz constant and observe that latent layers of adversarially trained models are still highly vulnerable to adversarial perturbations.
    
    \item We propose a Latent Adversarial Training (\textbf{LAT}) technique that significantly increases the robustness of existing state of the art adversarially trained models\cite{attack2017pgd} for MNIST, CIFAR-10, CIFAR-100, SVHN and Restricted ImageNet datasets. 

    \item We propose Latent Attack (\textbf{LA}), a new $l_{\infty}$ adversarial attack that is comparable in terms of performance to PGD on multiple datasets. The attack exploits the non-robustness of in-between layers of existing robust models to construct adversarial perturbations.
\end{itemize}

The rest of the paper is structured as follows: In Section \ref{related_work}, we review various adversarial attack and defense methodologies. In Section \ref{Section4} and \ref{lat}, we analyze the vulnerability of latent layers in robust models and introduce our proposed training technique of Latent Adversarial Training (LAT). In Section \ref{la} we describe our adversarial attack algorithm Latent Attack (LA). Further, we do some ablation studies to understand the effect of the choice of the layer on LAT and LA attack in Section \ref{discussion}. 


\section{Background And Related Work}
\label{related_work}
\subsection{Adversarial Attacks}
For a classification network $f$, let $\theta$ be its parameters, $y$ be the true class of $n$ - dimensional input $x \in [0, 1]^n$ and $J(\theta,x,y)$ be the loss function. The aim of an adversarial attack is to find the minimum perturbation $\Delta$ in $x$ that results in the change of class prediction. Formally, 
\begin{equation}
    \begin{aligned}
    &\Delta(x,f) := min_{\delta} ||\delta||_{p} \\ 
    \textnormal{s.t } &\arg\max(f(x+\delta;\theta)) \neq \arg\max(f(x;\theta))
    \end{aligned}
\end{equation}
It can be expressed as an optimization problem as: 
\begin{align*}
    x^{adv} = \underset{{\Tilde{x}:||\Tilde{x}-x||_p < \epsilon }}{\arg\max} J(\theta, \Tilde{x}, y)  
\end{align*}

In general, the magnitude of adversarial perturbation is constrained by a $p$ norm where $p\in\{0,2,\infty\}$ to ensure that the perturbed example is close to the original sample. Various other constraints for closeness and visual similarity \cite{stadv} have also been proposed for the construction of adversarial perturbation .\\
There are broadly two type of adversarial attacks:- White box and Black box attacks. White box attacks assume complete access to the network parameters while in the latter there is no information available about network architecture or parameters. We briefly describe PGD\cite{attack2017pgd} adversarial attack which we use as a baseline in our paper.

\paragraph{Projected Gradient Descent (PGD) attack }
Projected gradient descent \cite{attack2017pgd} is an iterative variant of Fast Gradient Sign Method (FGSM)\cite{goodfellow2014explaining}. Adversarial examples are constructed by iteratively applying FGSM and projecting the perturbed output to a valid constrained space $S$. 
The attack formulation is as follows:
\begin{equation}
    x^{i+1} = Proj_{x+S} \;(x^i  + \alpha \: sign(\nabla_{x} J(\theta,x^i,y)))
\end{equation}
where $x^{i+1}$ denotes the perturbed sample at $(i+1)_{th}$ iteration. 

While there has been extensive work in this area\cite{reviewpaper,reviewpaper2}, we primarily focus our attention towards attacks which utilizes latent layer representation. \cite{attack2017latent} proposed a method to construct adversarial perturbation by manipulating the latent layer of different classes. 
However, Latent Attack (LA) exploits the adversarial vulnerability of the latent layers to compute adversarial perturbations.



\subsection{Adversarial Defense } 
Popular defense strategies to improve the robustness of deep networks include the use of regularizers inspired by reducing the Lipschitz constant of the neural network \cite{tsuzuku2018lipschitz,cisse2017parseval}. There have also been several methods which turn to GAN's\cite{defensegan} for classifying the input as an adversary. However, these defense techniques were shown to be ineffective to adaptive adversarial attacks \cite{obfuscate,break_alp}. Hence we turn to adversarial training which \cite{goodfellow2014explaining,attack2017pgd,alp} is a defense technique that injects adversarial examples in the training batch at every step of the training. Adversarial training constitutes the current state-of-the-art in adversarial robustness against white-box attacks. For a comprehensive review of the work done in the area of adversarial examples, please refer \cite{reviewpaper,reviewpaper2}.

In our current work, we try to enhance the robustness of each latent layer, and hence increasing the robustness of the network as a whole. Previous works in this area include \cite{closestlayerwise,fb_noise}. However, our paper is different from them on the following counts: 
\begin{itemize}
\item \cite{fb_noise} observes that the adversarial perturbation on image leads to noisy features in latent layers. Inspired by this observation, they develop a new network architecture that comprises of denoising blocks at the feature layer which aims at increasing the adversarial robustness. However, we are leveraging the observation of low robustness at feature layer to perform adversarial training for latent layers to achieve higher robustness.
\item \cite{closestlayerwise} proposes an approach to regularize deep neural networks by perturbing intermediate layer activation. Their work has shown improvement in test accuracy over image classification tasks as well as minor improvement in adversarial robustness with respect to basic adversarial perturbation \cite{goodfellow2014explaining}. However, our work focuses on the vulnerability of latent layers to a small magnitude of adversarial perturbations. We have shown improvement in test accuracy and adversarial robustness with respect of state of the art attack \cite{attack2017pgd}.
\end{itemize}

\section{Robustness of Latent Layers}\label{Section4}

Mathematically, a deep neural network with $l$ layers and $f(x)$ as output can be described as:
\begin{equation}\label{eq1}
    f(x) = f_l(f_{l-1}(...(f_2(f_1(x;W_1,b_1);W_2,b_2)))...;W_l,b_l)
\end{equation}
Here $f_i$ denotes the function mapping layer $i-1$ to layer $i$ with weights $W_i$ and bias $b_i$ respectively. From Eq. \ref{eq1}, it is evident that $f(x)$ can be written as a composition of two functions:
\begin{equation}\label{eq2}
    \begin{split}
    f(x) &= g_i\circ h_{i}(x)\; | \; 0 \leq i \leq l-1 \\
    \textnormal{where } &\;f_0 = I\;\textnormal{and}\; h_i = f_i \circ f_{i-1} ... \circ f_1\circ f_0\\
    &g_i = f_l \circ f_{l-1} ... \circ f_{i+1} 
    \end{split} 
\end{equation}
We can study the behavior of $f(x)$ at a slightly perturbed input by inspecting its Lipschitz constant, which is defined by a constant $L_f$ such that Eq. \ref{eq_lipschitz} holds for all $\nu$. 
\begin{equation}\label{eq_lipschitz}
    ||f(x+\nu) - f(x) || \leq L_f||\nu||
\end{equation}
Having a lower Lipschitz constant ensures that function's output at perturbed input is not significantly different. This further can be  translated to higher adversarial robustness as it has been shown by \cite{cisse2017parseval,tsuzuku2018lipschitz}. Moreover, if  $L_g$ and $L_h$ are the Lipschitz constant of the sub-networks $g_i$ and $h_i$, the Lipschitz constant of $f$ has an upper bound defined by the product of Lipschitz constant of $g_i$ and $h_i$, i.e. 
\begin{equation}
    L_f \leq L_g*L_h
\end{equation}
So having robust sub-networks can result in higher adversarial robustness for the whole network $f$. But the converse need not be true.

For each of the latent layers $i$, we calculate an upper bound for the magnitude of perturbation($\epsilon_i$) by observing the perturbation induced in latent layer for adversarial examples $x^{adv}$.For obtaining a sensible bound of the perturbation for the sub-network $g_i(x)$, the following formula is used :
\begin{equation}\label{eq8}
    \epsilon_i \propto Mean_{x\in test}||h_i(x) - h_i(x^{adv})||_{\infty}
\end{equation}
Using this we compute the adversarial robustness of sub-networks $\{g_i | 1 \leq i \leq l-1 \}$ using PGD attack as shown in Fig \ref{latent_acc}.\\

\begin{figure}[t]
\setlength{\belowcaptionskip}{-10pt}
\begin{center}
    \includegraphics[width=3.0in,height=1.6in]{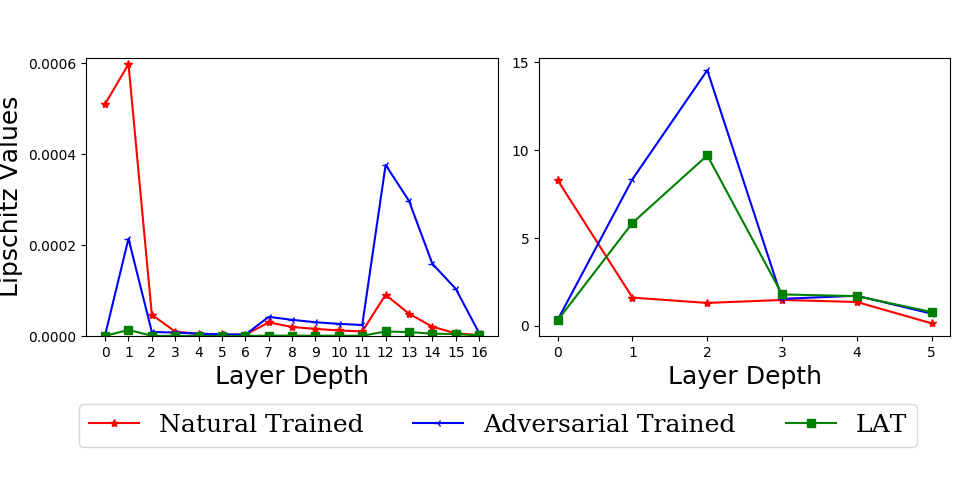} 
     \label{layer_cifar10_mnist_Adv}
    \caption{Lipschitz value of sub-networks with varying depth for different models on CIFAR-10 and MNIST}
    \label{latent_lipschitz}
\end{center}
\end{figure}

We now, briefly describe the network architecture used for each dataset.
\footnote{Code avaiable at: \url{https://github.com/msingh27/LAT_adversarial_robustness}}
\begin{itemize}
\item \textbf{MNIST}\cite{mnist_dataset}: We use the network architecture as described in \cite{attack2017pgd}. The natural and adversarial trained model achieves a test accuracy of 99.17\% and 98.4\% respectively.

\item \textbf{CIFAR-10}\cite{krizhevsky2010cifar}: We use the network architecture as in \cite{attack2017pgd}. The natural and adversarial trained model achieves a test accuracy of 95.01\% and 87.25\% respectively.

\item \textbf{CIFAR-100}\cite{krizhevsky2010cifar}: We use the same network architecture as used for CIFAR-10 with the modification at the logit layer so that it can handle the number of classes in CIFAR-100. The natural and adversarial trained model achieves a test accuracy of 78.07\% and 60.38\% respectively.

\item \textbf{SVHN}\cite{netzer2011reading}: We use the same network architecture as used for CIFAR-10. The adversarial trained model achieves a test accuracy of 91.13\%.

\item \textbf{Restricted Imagenet}\cite{tsipras2018robustness}: The dataset consists of a subset of imagenet classes which have been grouped into 9 different classes.
The model achieves a test accuracy of 91.65\%.
\end{itemize}

For adversarial training, the examples are constructed using PGD adversarial perturbations\cite{attack2017pgd}.  Also, we refer adversarial accuracy of a model as the accuracy over the adversarial examples generated using the test-set of the dataset. Higher adversarial accuracy corresponds to a more adversarially robust model.

We observe that for adversarially trained models, the adversarial accuracies of the sub-networks $g_{i}$ are relatively less than that of the whole network $f$ as shown in Fig \ref{latent_acc} and \ref{cifar100_lipschitz_adv}. The trend is consistent across all the different datasets.
Note that layer depth, i.e. $i$ is relative in all the experiments and the sampled layers are distributed evenly across the model. Also, in all tests the deepest layer tested is the layer just before the logit layer. Layer 0 corresponds to the input layer of $f$.

Fig \ref{latent_acc} and \ref{cifar100_lipschitz_adv} reveal that the sub-networks of an adversarially trained model are still vulnerable to adversarial perturbations. In general, it reduces with increasing depth. Though, a peculiar trend to observe is the increased robustness in the later layers of the network. The plots indicate that there is a scope of improvement in the adversarial robustness of different sub-networks. In the next section, we introduce our method that specifically targets at making $g_i$ robust. We find that this leads to a boost in the adversarial and test performance of the whole network $f$ as well.


To better understand the characteristics of sub-networks we do further analysis from the viewpoint of Lipschitz constant of the sub-networks. Since we are only concerned with the behavior of the function in the small neighborhood of input samples, we compute Lipschitz constant of the whole network $f$ and sub-networks $g_i$ using the local neighborhood of input samples i.e. 
\begin{equation}
    L_f(x_i) = max_{x_j \in B_{\epsilon}(x_i) }\dfrac{||f(x_j)-f(x_i)||}{||x_j-x_i||}
\end{equation}
where $B_{\epsilon}(x_i)$ denotes the $\epsilon$ neighbourhood of $x_i$. For computational reasons, inspired by \cite{alvarez2018robustness}, we approximate $B_{\epsilon}(x_i)$ by adding noise to $x_i$ with epsilon as given in Eq. \ref{eq8}. We report the value averaged over complete test data for different datasets and models in Fig. \ref{latent_lipschitz}. 
The plot reveals that while for the adversarially trained model, the Lipschitz value of $f$ is lower than that of the naturally trained model, there is no such pattern in the sub-networks $g_i$. This observation again reinforces our hypothesis of the vulnerabilities of the different sub-networks against small perturbations.

\section{Harnessing Latent Layers}

\subsection{Latent Adversarial Training (LAT)}\label{lat}

In this section, we seek to increase the robustness of the deep neural network, $f$. We propose Latent Adversarial training (LAT) wherein both $f$ and one of the sub-networks $g_i$ are adversarially trained. For adversarial training of $g_i$, we use a $l_{\infty}$ bounded adversarial perturbation computed via the PGD attack at layer $i$ with appropriate bound as defined in Eq. \ref{eq8}. 


We are using LAT as a fine-tuning technique which operates on a adversarially trained model to improve its adversarial and test accuracy further. We observe that performing only a few epochs ($\sim 5$) of LAT on the adversarially trained model results in a significant improvement over adversarial accuracy of the model. Algorithm \ref{feature_adv_train_algo} describes our LAT training technique. 

To test the efficacy of LAT, we perform experiments over CIFAR-10, CIFAR-100, SVHN, Rest. Imagenet and MNIST datasets. For fairness, we also compare our approach (LAT) against two baseline fine-tuning techniques.
\begin{itemize}
    \item Adversarial Training (AT) \cite{attack2017pgd}
    \item Feature Noise Training (FNT) using algorithm \ref{feature_adv_train_algo} with gaussian noise to perturb the latent layer $i$.
\end{itemize}

Table \ref{regularize_all} reports the adversarial accuracy corresponding to LAT and baseline fine-tuning methods over the different datatsets. PGD Baseline corresponds to 10 steps for CIFAR-10, CIFAR-100 and SVHN, 40 steps for MNIST and 8 steps of PGD attack for Restricted Imagenet.
We perform 2 epochs of fine-tuning for MNIST, CIFAR-10, Rest. Imagenet, 1 epoch for SVHN and 5 epochs for CIFAR-100 using the different techniques. The results are calculated with the constraint on the maximum amount of per-pixel perturbation as $0.3/1.0$ for MNIST dataset and $8.0/255.0$ for CIFAR-10, CIFAR-100, Restricted ImageNet and SVHN.

\setlength{\textfloatsep}{0.1cm}
\setlength{\floatsep}{0.1cm}
\begin{algorithm}[H]
\SetAlgoLined
\Begin{
\footnotesize{
    \textbf{Input}: Adversarially trained model parameters - $\theta$, Sub-network index which needs to be adversarially trained - $m$, Fine-tuning steps - $k$, Batch size - $B$, Learning rate - $\eta$, hyperparameter $\omega$ \\
    \textbf{Output}: Fine-tuned model parameters\\ 
    
    \For{$i \in {1, 2, ..., k}$}
    {
        
        Training data of size B - $(X(i), Y(i))$. \\
        Compute adversarial perturbation $\Delta{X(i)}$ via PGD attack.\\
        Calculate the gradients $J_{adv} = J(\theta, X(i) + \Delta{X(i)}, Y(i))$. \\
        Compute $h_m(X(i))$.\\
        Compute $\epsilon$ corresponding to $(X(i), Y(i))$ via Eq. \ref{eq8}.\\
        Compute adversarial perturbation $\Delta{h_m(X(i))}$ with perturbation amount $\epsilon$ \\
        Compute the gradients $J_{latentAdv} = J(\theta, h_m(X(i)) + \Delta{h_m(X(i))}, Y(i))$\\
        
        $J(\theta, X(i), Y(i)) =  \omega * J_{adv} + (1 - \omega) * (J_{latentAdv})$\\
        
        $\theta \rightarrow \theta - \eta * J(\theta, X(i), Y(i)) $

    }
    \textbf{return} fine-tuned model.
    }
 }
 \caption{\footnotesize{Algorithm for improving the adversarial robustness of models}}
\label{feature_adv_train_algo}
\end{algorithm}
\setlength{\textfloatsep}{0.08cm}
\setlength{\floatsep}{0.08cm}












\begin{table}[t]
\begin{center}
\footnotesize{
\begin{tabular}{ll|l|l|l}
\cline{3-4}& & \multicolumn{2}{l|}{Adversarial Accuracy} & \\ \hline
\multicolumn{1}{|l|}{Dataset}                    & \begin{tabular}[c]{@{}l@{}}Fine-tune\\ Technique\end{tabular} & \begin{tabular}[c]{@{}l@{}}PGD  \\ Baseline\end{tabular} & \begin{tabular}[c]{@{}l@{}}PGD\\ (100 step)\end{tabular}  & \multicolumn{1}{l|}{Test Acc.}         \\ \hline
\multicolumn{1}{|l|}{\multirow{3}{*}{CIFAR-10}}  & AT                                                              & 47.12 \%             & 46.19 \%           & \multicolumn{1}{l|}{87.27 \%}          \\ \cline{2-5} 
\multicolumn{1}{|l|}{}                           & FNT                                                             & 46.99 \%             & 46.41 \%           & \multicolumn{1}{l|}{87.31 \%}          \\ \cline{2-5} 
\multicolumn{1}{|l|}{}                           & LAT                                                             & \textbf{53.84 \%}    & \textbf{53.04 \%}           & \multicolumn{1}{l|}{\textbf{87.80 \%}} \\ \hline

\multicolumn{1}{|l|}{\multirow{3}{*}{CIFAR-100}} & AT                                                              & 22.72 \%             & 22.21 \%           & \multicolumn{1}{l|}{60.38 \%}          \\ \cline{2-5} 
\multicolumn{1}{|l|}{}                           & FNT                                                            & 22.44 \%             & 21.86 \%           & \multicolumn{1}{l|}{60.27 \%}          \\ \cline{2-5} 
\multicolumn{1}{|l|}{}                           & LAT                                                             & \textbf{27.03 \%}    & \textbf{26.41 \%}           & \multicolumn{1}{l|}{\textbf{60.94 \%}} \\ \hline

\multicolumn{1}{|l|}{\multirow{3}{*}{SVHN}} & AT                                                              & 54.58 \%             & 53.52 \%           & \multicolumn{1}{l|}{91.88 \%}          \\ \cline{2-5} 
\multicolumn{1}{|l|}{}                           & FNT                                                            & 54.69 \%             & 53.96 \%           & \multicolumn{1}{l|}{\textbf{92.45 \%}}          \\ \cline{2-5} 
\multicolumn{1}{|l|}{}                           & LAT                                                             & \textbf{60.23 \%}    & \textbf{59.97 \%}           & \multicolumn{1}{l|}{91.65 \%} \\ \hline

\multicolumn{1}{|l|}{{Rest. }}& AT & 17.52 \%& 16.04\% & \multicolumn{1}{l|}{\textbf{91.83 \%}} \\ \cline{2-5}  \multicolumn{1}{|l|}{ImageNet} & FNT                                                             & 18.81 \%             & 17.32 \%           & \multicolumn{1}{l|}{91.59 \%}          \\ \cline{2-5} 
\multicolumn{1}{|l|}{}                           & LAT                                                             & \textbf{22.00 \%}    & \textbf{20.11 \%}           & \multicolumn{1}{l|}{89.86 \%}          \\ \hline

\multicolumn{1}{|l|}{\multirow{3}{*}{MNIST}}     & AT                                                              & 93.75 \%             & 92.92\%         & \multicolumn{1}{l|}{\textbf{98.40 \%}} \\ \cline{2-5} 
\multicolumn{1}{|l|}{}                           & FNT                                                             & 93.59 \%             & 92.16 \%           & \multicolumn{1}{l|}{98.28 \%}          \\ \cline{2-5} 
\multicolumn{1}{|l|}{}                           & LAT                                                             & \textbf{94.21 \%}    & \textbf{93.31 \%}           & \multicolumn{1}{l|}{98.38 \%}          \\ \hline

\end{tabular}
\caption{\footnotesize{Adversarial accuracy for different datasets after fine-tuning using different methods} }
\label{regularize_all}
}
\end{center}
\end{table}







The results in the Table \ref{regularize_all} correspond to the best performing layers \footnote{The results correspond to $g_{11}$, $g_{10}$, $g_{7}$, $g_7$ and $g_2$ sub-networks for the CIFAR-10, SVHN, CIFAR-100, Rest. Imagenet and MNIST datasets respectively.}.
As can be seen from the table, that only after 2 epochs of training by LAT on CIFAR-10 dataset, the adversarial accuracy jumps by $\sim 6.5\%$. Importantly, LAT not only improves the performance of the model over the adversarial examples but also over the clean test samples, which is reflected by an improvement of 0.6\% in test accuracy. A similar trend is visible for SVHN and CIFAR-100 datasets where LAT improves the adversarial accuracy by 8\% and 4\% respectively, as well as the test accuracy for CIFAR-100 by 0.6\% .
 Table \ref{regularize_all} also reveals that the two baseline methods do not lead to any significant changes in the performance of the model. As the adversarial accuracy of the adversarially trained model for the MNIST dataset is already high (93.75\%), our approach does not lead to significant improvements ($\sim 0.46\%$).

\begin{figure}[t]
\begin{center}
    \includegraphics[width=3.5in,height=1.7in]{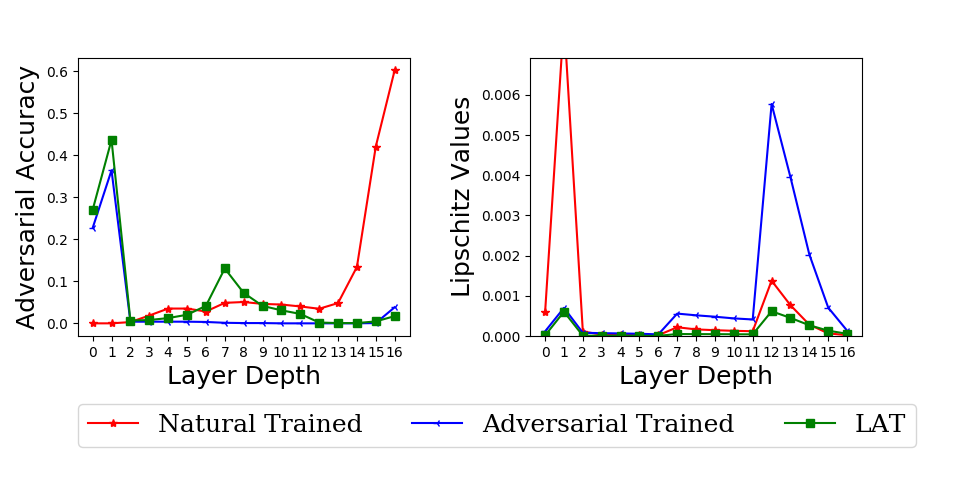} 
     \label{layer_cifar100}
    \caption{\footnotesize{Adversarial accuracy and Lipschitz values with varying depth for different models on CIFAR-100 }}
    \label{cifar100_lipschitz_adv}
\end{center}
\end{figure}
To analyze the effect of LAT on latent layers, we compute the robustness of various sub-networks $g_i$ of $f$ after training using LAT. Fig \ref{latent_acc} shows the robustness of different sub-network $g_i$ with and without our LAT method for CIFAR-10 and MNIST datasets. Figure \ref{cifar100_lipschitz_adv} contains the results for CIFAR-100 dataset.
As the plots show, our approach not only improves the robustness of $f$ but also that of most of the sub-networks $g_i$.
A detailed analysis analyzing the effect of the choice of the layer and the hyperparameter $\omega$ of LAT on the adversarial robustness of the model is shown in section \ref{discussion}.
    

    





    

\subsection{Latent Adversarial Attack (LA)}\label{la}
In this section, we seek to leverage the vulnerability of latent layers of a neural network to construct adversarial perturbations. In general, existing adversarial perturbation calculation methods like FGSM \cite{goodfellow2014explaining} and PGD \cite{attack2017pgd} operate by directly perturbing the input layer to optimize the objective that promotes misclassification. In our approach, for given input example $x$ and a sub-network $g_i(x)$, we first calculate adversarial perturbation $\Delta(x,g_i)$ constrained by appropriate bounds where $i\in (1,2,..,l)$.
Here,
\begin{equation}
\begin{aligned}
& \Delta(x,g_i) := \underset{\delta}{\min} ||\delta||_{p} \:\; where\:\;  p\in \{2,\infty\} \\
& \textnormal{s.t}\;\; \arg \max(g_i(h_i(x)+\delta)) \neq \arg\max(g_i(h_i(x)))
\end{aligned}
\end{equation}
Subsequently, we optimize the following equation to obtain $\Delta(x,f)$ for LA : 
\begin{equation}
 \Delta(x,f) = \underset{\mu}{\arg \min} |h(x+\mu) - (h(x) + \Delta(x,g_i)| 
\end{equation}
We repeat the above two optimization steps iteratively to obtain our adversarial perturbation.

For the comparison of the performance of LA, we use PGD adversarial perturbation as a baseline attack. In general, we obtain better or comparable adversarial accuracy when compared to PGD attack. We use the same configuration for $\epsilon$ as in LAT. 
For MNIST and CIFAR-100, our LA achieves an adversarial accuracy of $90.78\%$ and $22.87\%$ respectively whereas PGD(100 steps) and PGD(10 steps) obtains adversarial accuracy of $92.52\%$ and $23.01\%$ respectively. In the case of CIFAR-10 dataset, LA achieves adversarial accuracy of $47.46\%$ and PGD(10 steps) obtains adversarial accuracy of $47.41$. The represented LA attacks are from the best layers, i.e., $g_1$ for MNIST, CIFAR-100 and $g_2$ for CIFAR-10.

\begin{figure}[t]
\begin{center}
    \includegraphics[width=3.5in,height=1.7in]{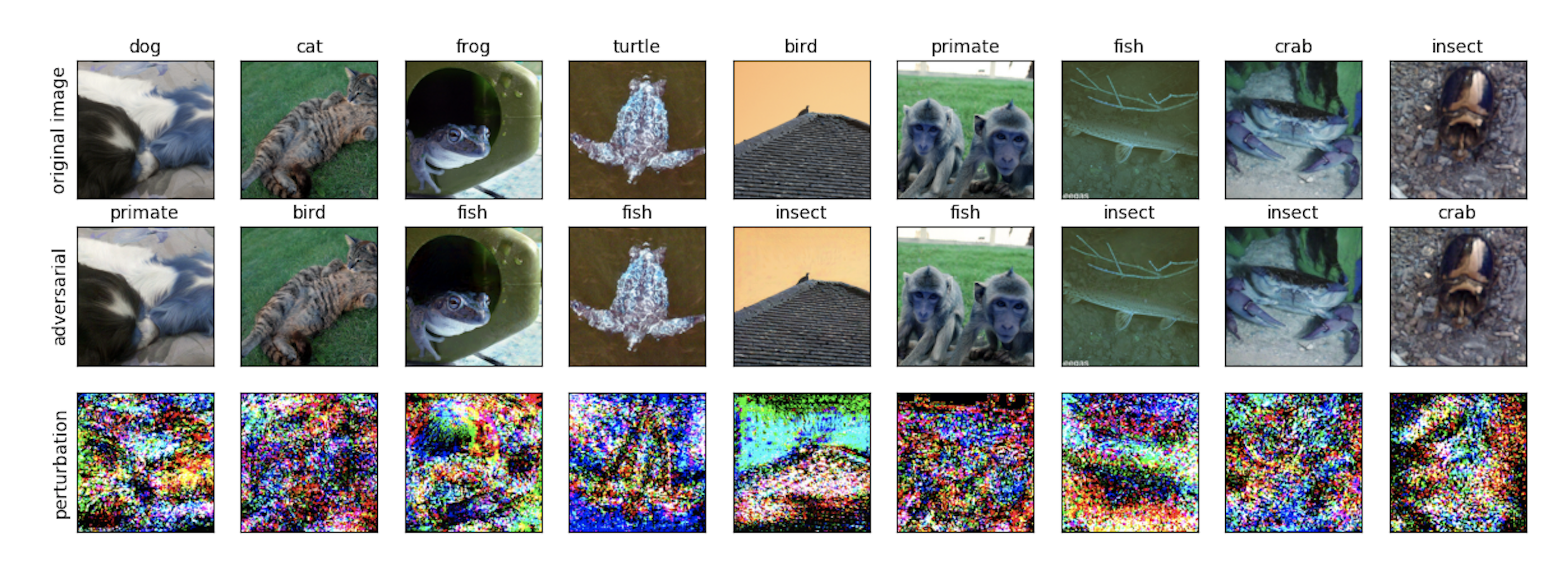} 
     \label{layer_adv_training_acc}
    \caption{\footnotesize{Adversarial images of Restricted ImageNet constructed using Latent Adversarial Attack (LA)}}
    \label{imagenet_la_image}
\end{center}
\end{figure}

Some of the adversarial examples generated using LA is illustrated in Fig \ref{imagenet_la_image}.
The pseudo code of the proposed algorithm(LA)is given in Algo $\ref{latent_adv_attack}$.

\setlength{\textfloatsep}{0.05cm}
\setlength{\floatsep}{0.05cm}
\begin{algorithm}[H]

\SetAlgoLined
\Begin{
\footnotesize{
    \textbf{Input}: Neural network model $f$, sub-network $g_m$, step-size for latent layer $\alpha_l$, step-size for input layer $\alpha_x$, intermediate iteration steps $p$, global iteration steps - $k$, input example $x$, adversarial perturbation generation technique for $g_m$
    \newline
    \textbf{Output}: Adversarial example 
    
    $x^{1}$ = $x$
    \For{$i \in {1, 2, ..., k}$}
    {   
        $l^{1}$ = $g_m(x^i)$
        \newline
        \For{$j \in {1, 2, ..., p}$}{
        $$l^{j+1} = Proj_{l+S} \;(l^j  + \alpha_l \: sign(\nabla_{g_m(x)} J(\theta,x,y))) $$
        }
        $x_{adv}^1$ = $x^i$
        \newline
        \For{$j \in {1, 2, ..., p}$}{
        $$x_{adv}^{j+1} = Proj_{x_{adv}+S}(x_{adv}^j  - \alpha_xsign(\nabla_{x} |g_m(x) - l^p| )) $$
        }
        $x^i$ = $x_{adv}^p$
    }
    \textbf{return} $x^{k}$
    }
 }
\caption{\footnotesize{Proposed algorithm for the construction of adversarial perturbation}}
\label{latent_adv_attack}
\end{algorithm}
\setlength{\textfloatsep}{0.075cm}
\setlength{\floatsep}{0.075cm}





\section{Discussion and Ablation Studies}\label{discussion}
\begin{figure}[t]
\begin{center}
    \includegraphics[width=1.6in,height=1.2in]{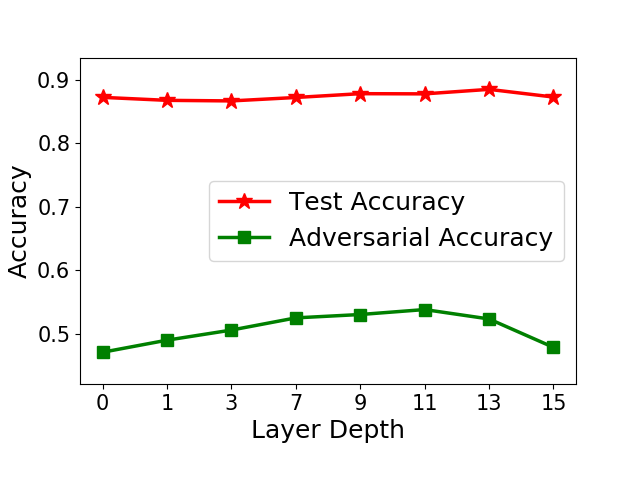} 
    \includegraphics[width=1.6in,height=1.2in]{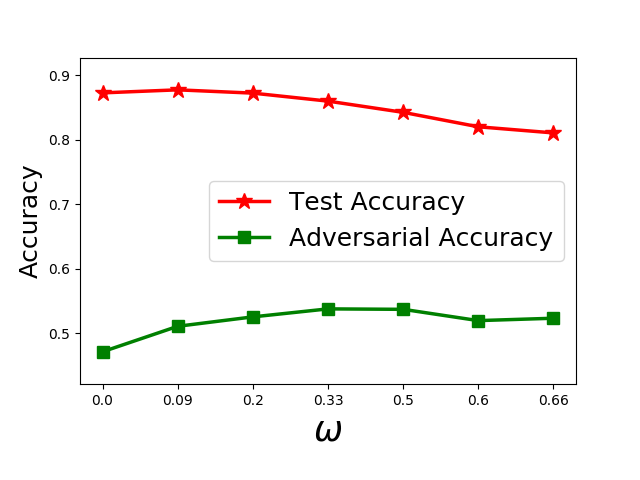} 
    \caption{\footnotesize{Plot showing effect of layer depth and $\omega$ on the adversarial and test accuracies of $f(x)$ on CIFAR-10}}
    \label{alpha_adv_training_acc}
\end{center}
\end{figure}

To gain an understanding of LAT, we perform various experiments and analyze the findings in this section. We choose CIFAR-10 as the primary dataset for all the following experiments.
\paragraph{Effect of layer depth in LAT.} We fix the value of $\omega$ to the best performing value of $0.2$ and fine-tune the model using LAT for different latent layers of the network. The left plot in Fig \ref{alpha_adv_training_acc} shows the influence of the layer depth in the performance of the model. It can be observed from the plot, that the robustness of $f$ increases with increasing layer depth, but the trend reverses for the later layers. This observation can be explained from the plot in Fig \ref{latent_acc}, where the robustness of $g_i$ decreases with increasing layer depth $i$, except for the last few layers.


\paragraph{Effect of hyperparameter $\omega$ in LAT.} We fix the layer depth to $11(g_{11})$ as it was the best performing layer for CIFAR-10 and we perform LAT for different values of $\omega$. This hyperparameter $\omega$ controls the ratio of weight assigned to the classification loss corresponding to adversarial examples for $g_{11}$ and the classification loss corresponding to adversarial examples for $f$. The right plot in Fig \ref{alpha_adv_training_acc} shows the result of this experiment. We find that the robustness of $f$ increases with increasing $\omega$. However, the adversarial accuracy does start to saturate after a certain value. The performance of test accuracy also starts to suffer beyond this point.

\begin{figure}[t]
\begin{center}
    \includegraphics[width=3.7in,height=1.5in]{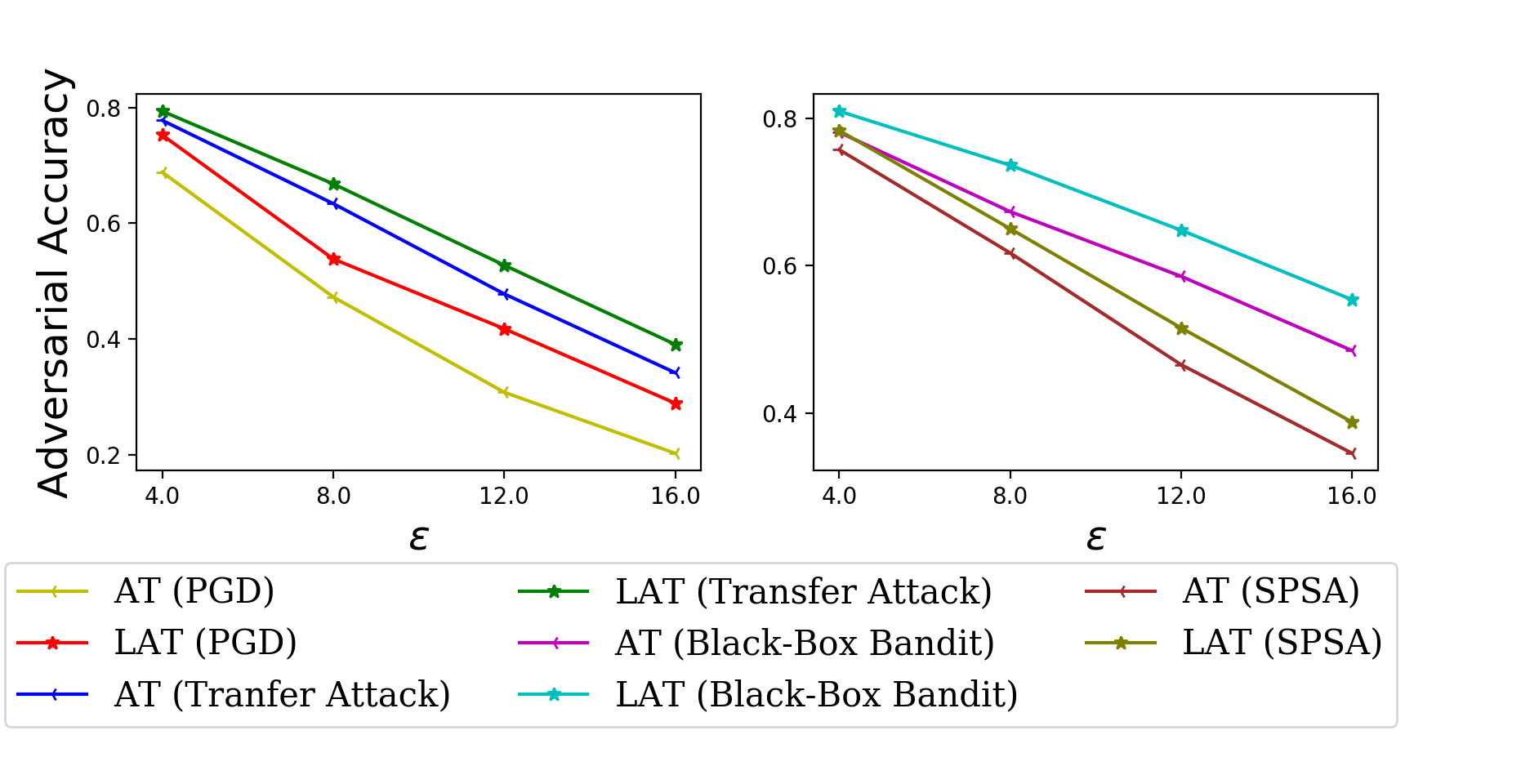} 
     \label{layer_black_box}
    \caption{\footnotesize{White-Box and Black-Box Adversarial accuracy on various $\epsilon$ on CIFAR-10}}
    \label{eps_attacks_acc}
\end{center}
\end{figure}

\paragraph{Black-box and white-box attack robustness.}
We test the black box and white-box adversarial robustness of LAT fine-tuned model for the CIFAR-10 dataset over various $\epsilon$ values. For evaluation in black box setting, we perform transfer attack from a secret adversarially trained model, bandit black box attack\cite{ilyas2018prior} and SPSA\cite{uesato2018adversarial}. Figure \ref{eps_attacks_acc} shows the adversarial accuracy. As it can be seen, the LAT trained model achieves higher adversarial robustness for both the black box and white-box attacks over a range of $\epsilon$ values when compared against baseline AT model.
We also observe that the adversarial perturbations transfers better({$\sim 1\%$}) from LAT model than AT models. 

\begin{figure}[t]
\begin{center}
    \includegraphics[width=3.4in,height=1.35in]{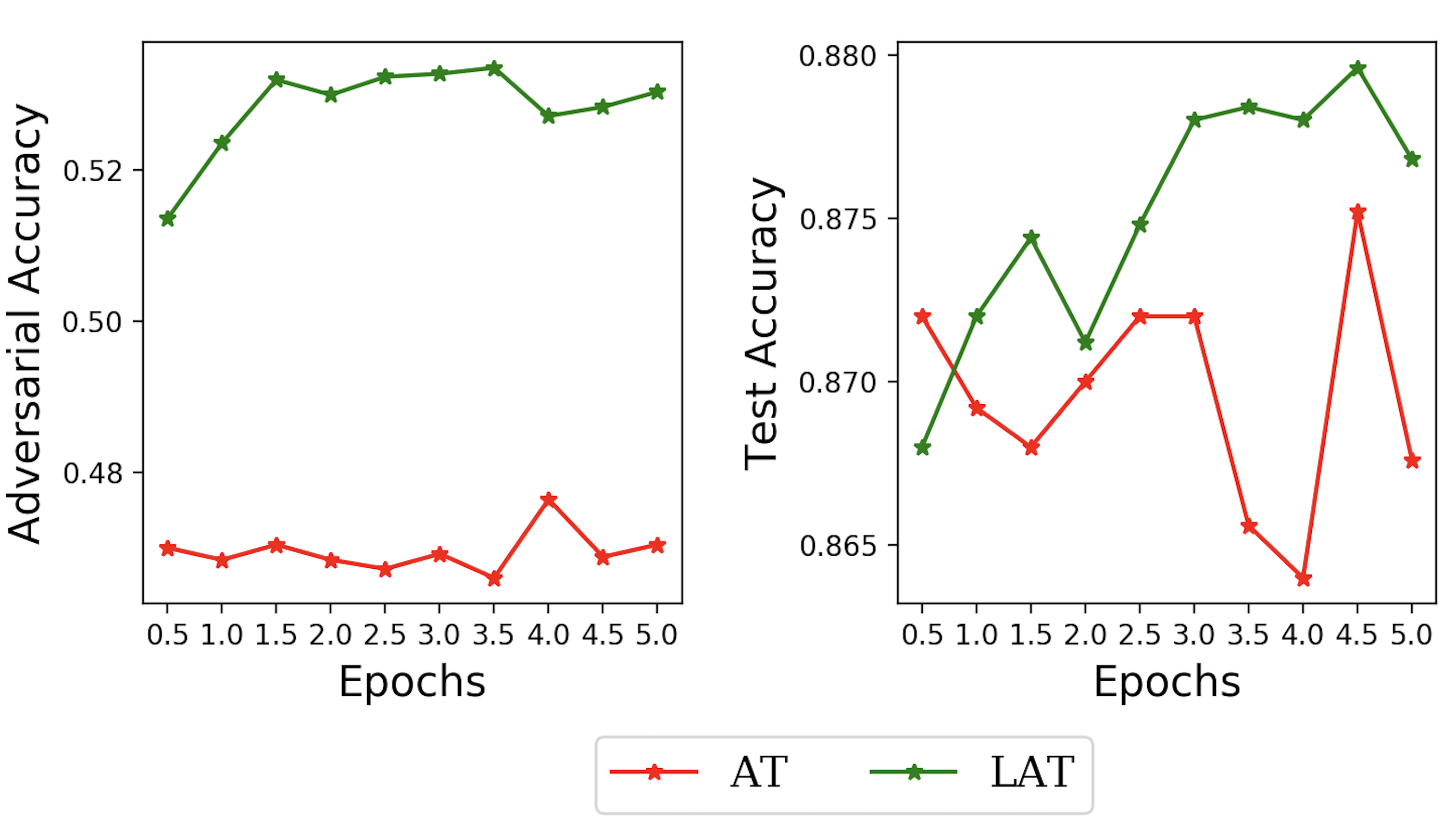} 
     \label{layer_training_time}
    \caption{\footnotesize{Progress of Adversarial and Test Accuracy for LAT and AT when fine-tuned for 5 epochs on CIFAR-10}}
    \label{training_time}
\end{center}
\end{figure}

\paragraph{Performance of LAT with training steps.} Figure \ref{training_time} plots the variation of test and adversarial accuracy while fine-tuning using the LAT and AT techniques.

\paragraph{Different attack methods used for LAT.}
Rather than using a $l_{\infty}$ bound PGD adversarial attack, we also explored using a $l_2$ bound PGD attack and FGSM attack to perturb the latent layers in LAT. By using $l_2$ bound PGD attack in LAT for 2.5 epochs, the model achieves an adversarial and test accuracy of \textbf{88.02\%} and \textbf{53.46\%} respectively. Using FGSM to perform LAT did not lead to improvement as the model achieves 48.83\% and 87.26\% adversarial and test accuracy respectively. 
The results are calculated by choosing the $g_{11}$ sub-network.

\paragraph{Random layer selection in LAT :}
Previous experiments of LAT fine-tuning corresponds to selecting a single sub-network $g_i$ and adversarially training it. We perform an experiment where at each training step of LAT we randomly choose one of the [$g_5$, $g_7$, $g_9$, $g_{11}$] sub-networks to perform adversarial training. The model performs comparably, achieving a test and adversarial accuracy of $87.31\%$ and $53.50\%$ respectively. 



\section{Conclusion}

We observe that deep neural network models trained via adversarial training have sub-networks vulnerable to adversarial perturbation. We described a latent adversarial training (LAT) technique aimed at improving the adversarial robustness of the sub-networks. We verified that using LAT significantly improved the adversarial robustness of the overall model for several different datasets along with an increment in test accuracy. We performed several experiments to analyze the effect of depth on LAT and showed higher robustness to Black-Box attacks. We proposed Latent Attack (LA) an adversarial attack algorithm that exploits the adversarial vulnerability of latent layer to construct adversarial examples. Our results show that the proposed methods that harness the effectiveness of latent layers in a neural network beat state-of-the-art in defense methods, and offer a significant pathway for new developments in adversarial machine learning.

\bibliographystyle{named}
\bibliography{ijcai19}

\begin{thebibliography}{}

\bibitem[\protect\citeauthoryear{Akhtar and Mian}{2018}]{reviewpaper2}
Naveed Akhtar and Ajmal Mian.
\newblock Threat of adversarial attacks on deep learning in computer vision: A
  survey.
\newblock {\em IEEE Access}, 2018.

\bibitem[\protect\citeauthoryear{Alvarez-Melis and
  Jaakkola}{2018}]{alvarez2018robustness}
David Alvarez-Melis and Tommi~S Jaakkola.
\newblock On the robustness of interpretability methods.
\newblock {\em ICML 2018 Workshop}, 2018.

\bibitem[\protect\citeauthoryear{Athalye \bgroup \em et al.\egroup
  }{2018}]{obfuscate}
Anish Athalye, Nicholas Carlini, and David Wagner.
\newblock Obfuscated gradients give a false sense of security: Circumventing
  defenses to adversarial examples.
\newblock {\em ICML}, 2018.

\bibitem[\protect\citeauthoryear{Carlini and
  Wagner}{2017a}]{carlini2017adversarial}
Nicholas Carlini and David Wagner.
\newblock Adversarial examples are not easily detected: Bypassing ten detection
  methods.
\newblock In {\em AISec}. ACM, 2017.

\bibitem[\protect\citeauthoryear{Carlini and
  Wagner}{2017b}]{carlini2017towards}
Nicholas Carlini and David Wagner.
\newblock Towards evaluating the robustness of neural networks.
\newblock In {\em 2017 IEEE Symposium on Security and Privacy (SP)}, 2017.

\bibitem[\protect\citeauthoryear{Carlini \bgroup \em et al.\egroup
  }{2017}]{provable_certificate4}
Nicholas Carlini, Guy Katz, Clark Barrett, and David~L Dill.
\newblock Provably minimally-distorted adversarial examples.
\newblock {\em arXiv preprint arXiv:1709.10207}, 2017.

\bibitem[\protect\citeauthoryear{Cihang~Xie}{2019}]{fb_noise}
Laurens van der Maaten Alan Yuille Kaiming~He Cihang~Xie, Yuxin~Wu.
\newblock Feature denoising for improving adversarial robustness.
\newblock {\em CVPR}, 2019.

\bibitem[\protect\citeauthoryear{Cisse \bgroup \em et al.\egroup
  }{2017}]{cisse2017parseval}
Moustapha Cisse, Piotr Bojanowski, Edouard Grave, Yann Dauphin, and Nicolas
  Usunier.
\newblock Parseval networks: Improving robustness to adversarial examples.
\newblock In {\em ICML}, 2017.

\bibitem[\protect\citeauthoryear{Goodfellow \bgroup \em et al.\egroup
  }{2015}]{goodfellow2014explaining}
Ian~J. Goodfellow, Jonathon Shlens, and Christian Szegedy.
\newblock Explaining and harnessing adversarial examples.
\newblock {\em ICLR}, 2015.

\bibitem[\protect\citeauthoryear{He \bgroup \em et al.\egroup
  }{2016}]{he2016deep}
Kaiming He, Xiangyu Zhang, Shaoqing Ren, and Jian Sun.
\newblock Deep residual learning for image recognition.
\newblock In {\em CVPR}, 2016.

\bibitem[\protect\citeauthoryear{Ilyas \bgroup \em et al.\egroup
  }{2019}]{ilyas2018prior}
Andrew Ilyas, Logan Engstrom, and Aleksander Madry.
\newblock Prior convictions: Black-box adversarial attacks with bandits and
  priors.
\newblock {\em ICLR}, 2019.

\bibitem[\protect\citeauthoryear{Kannan \bgroup \em et al.\egroup }{2018}]{alp}
Harini Kannan, Alexey Kurakin, and Ian~J. Goodfellow.
\newblock Adversarial logit pairing.
\newblock {\em NIPS}, 2018.

\bibitem[\protect\citeauthoryear{Krizhevsky \bgroup \em et al.\egroup
  }{2010}]{krizhevsky2010cifar}
Alex Krizhevsky, Vinod Nair, and Geoffrey Hinton.
\newblock Cifar-10.
\newblock {\em URL \url{http://www. cs. toronto. edu/kriz/cifar. html}}, 2010.

\bibitem[\protect\citeauthoryear{Krizhevsky \bgroup \em et al.\egroup
  }{2012}]{krizhevsky2012imagenet}
Alex Krizhevsky, Ilya Sutskever, and Geoffrey~E Hinton.
\newblock Imagenet classification with deep convolutional neural networks.
\newblock In {\em NIPS}, 2012.

\bibitem[\protect\citeauthoryear{Lecun \bgroup \em et al.\egroup
  }{1989}]{mnist_dataset}
Yan Lecun, B.~Boser, J.S. Denker, D.~Henderson, R.E. Howard, W.~Hubbard, and
  L.D. Jackel.
\newblock Backpropagation applied to handwritten zip code recognition, 1989.

\bibitem[\protect\citeauthoryear{Logan~Engstrom}{2018}]{break_alp}
Anish~Athalye Logan~Engstrom, Andrew~Ilyas.
\newblock Evaluating and understanding the robustness of adversarial logit
  pairing.
\newblock {\em NeurIPS SECML}, 2018.

\bibitem[\protect\citeauthoryear{Madry \bgroup \em et al.\egroup
  }{2018}]{attack2017pgd}
Aleksander Madry, Aleksandar Makelov, Ludwig Schmidt, Dimitris Tsipras, and
  Adrian Vladu.
\newblock Towards deep learning models resistant to adversarial attacks.
\newblock {\em ICLR}, 2018.

\bibitem[\protect\citeauthoryear{Moosavi-Dezfooli \bgroup \em et al.\egroup
  }{2017}]{moosavi2017universal}
Seyed-Mohsen Moosavi-Dezfooli, Alhussein Fawzi, Omar Fawzi, and Pascal
  Frossard.
\newblock Universal adversarial perturbations.
\newblock {\em CVPR}, 2017.

\bibitem[\protect\citeauthoryear{Netzer \bgroup \em et al.\egroup
  }{2011}]{netzer2011reading}
Yuval Netzer, Tao Wang, Adam Coates, Alessandro Bissacco, Bo~Wu, and Andrew~Y
  Ng.
\newblock Reading digits in natural images with unsupervised feature learning.
\newblock {\em NIPS Workshop}, 2011.

\bibitem[\protect\citeauthoryear{Raghunathan \bgroup \em et al.\egroup
  }{2018}]{raghunathan2018certified}
Aditi Raghunathan, Jacob Steinhardt, and Percy Liang.
\newblock Certified defenses against adversarial examples.
\newblock {\em ICLR}, 2018.

\bibitem[\protect\citeauthoryear{Russakovsky \bgroup \em et al.\egroup
  }{2015}]{russakovsky2015imagenet}
Olga Russakovsky, Jia Deng, Hao Su, Jonathan Krause, Sanjeev Satheesh, Sean Ma,
  Zhiheng Huang, Andrej Karpathy, Aditya Khosla, Michael Bernstein, et~al.
\newblock Imagenet large scale visual recognition challenge.
\newblock {\em IJCV}, 2015.

\bibitem[\protect\citeauthoryear{Sabour \bgroup \em et al.\egroup
  }{2016}]{attack2017latent}
Sara Sabour, Yanshuai Cao, Fartash Faghri, and David~J. Fleet.
\newblock Adversarial manipulation of deep representations.
\newblock {\em ICLR}, 2016.

\bibitem[\protect\citeauthoryear{Samangouei \bgroup \em et al.\egroup
  }{2018}]{defensegan}
Pouya Samangouei, Maya Kabkab, and Rama Chellappa.
\newblock Defense-gan: Protecting classifiers against adversarial attacks using
  generative models.
\newblock {\em ICLR}, 2018.

\bibitem[\protect\citeauthoryear{Sankaranarayanan \bgroup \em et al.\egroup
  }{2018}]{closestlayerwise}
Swami Sankaranarayanan, Arpit Jain, Rama Chellappa, and Ser~Nam Lim.
\newblock Regularizing deep networks using efficient layerwise adversarial
  training.
\newblock {\em AAAI}, 2018.

\bibitem[\protect\citeauthoryear{Szegedy \bgroup \em et al.\egroup
  }{2014}]{szegedy2013intriguing}
Christian Szegedy, Wojciech Zaremba, Ilya Sutskever, Joan Bruna, Dumitru Erhan,
  Ian~J. Goodfellow, and Rob Fergus.
\newblock Intriguing properties of neural networks.
\newblock {\em ICLR}, 2014.

\bibitem[\protect\citeauthoryear{Tram{\`e}r \bgroup \em et al.\egroup
  }{2018}]{tramer2017ensemble}
Florian Tram{\`e}r, Alexey Kurakin, Nicolas Papernot, Ian~J. Goodfellow, Dan
  Boneh, and Patrick McDaniel.
\newblock Ensemble adversarial training: Attacks and defenses.
\newblock {\em ICLR}, 2018.

\bibitem[\protect\citeauthoryear{Tsipras \bgroup \em et al.\egroup
  }{2019}]{tsipras2018robustness}
Dimitris Tsipras, Shibani Santurkar, Logan Engstrom, Alexander Turner, and
  Aleksander Madry.
\newblock Robustness may be at odds with accuracy.
\newblock {\em ICLR}, 2019.

\bibitem[\protect\citeauthoryear{Tsuzuku \bgroup \em et al.\egroup
  }{2018}]{tsuzuku2018lipschitz}
Yusuke Tsuzuku, Issei Sato, and Masashi Sugiyama.
\newblock Lipschitz-margin training: Scalable certification of perturbation
  invariance for deep neural networks.
\newblock {\em NeurIPS}, 2018.

\bibitem[\protect\citeauthoryear{Uesato \bgroup \em et al.\egroup
  }{2018}]{uesato2018adversarial}
Jonathan Uesato, Brendan O'Donoghue, Aaron van~den Oord, and Pushmeet Kohli.
\newblock Adversarial risk and the dangers of evaluating against weak attacks.
\newblock {\em ICML}, 2018.

\bibitem[\protect\citeauthoryear{Weng \bgroup \em et al.\egroup
  }{2018}]{weng2018towards}
Tsui-Wei Weng, Huan Zhang, Hongge Chen, Zhao Song, Cho-Jui Hsieh, Duane Boning,
  Inderjit~S Dhillon, and Luca Daniel.
\newblock Towards fast computation of certified robustness for relu networks.
\newblock {\em ICML}, 2018.

\bibitem[\protect\citeauthoryear{Wong and Kolter}{2018}]{wong2017provable}
Eric Wong and J~Zico Kolter.
\newblock Provable defenses against adversarial examples via the convex outer
  adversarial polytope.
\newblock {\em ICML}, 2018.

\bibitem[\protect\citeauthoryear{Xiao \bgroup \em et al.\egroup }{2018}]{stadv}
Chaowei Xiao, Jun-Yan Zhu, Bo~Li, Warren He, Mingyan Liu, and Dawn Song.
\newblock Spatially transformed adversarial examples.
\newblock {\em ICLR}, 2018.

\bibitem[\protect\citeauthoryear{Yuan \bgroup \em et al.\egroup
  }{2019}]{reviewpaper}
Xiaoyong Yuan, Pan He, Qile Zhu, and Xiaolin Li.
\newblock Adversarial examples: Attacks and defenses for deep learning.
\newblock {\em IEEE transactions on neural networks and learning systems},
  2019.

\end{thebibliography}

\end{document}